# Small Business Classification By Name: Addressing Gender and Geographic Origin Biases

Daniel Shapiro

Lemay.ai

**Abstract.** Small business classification is a difficult and important task within many applications, including customer segmentation. Training on small business names introduces gender and geographic origin biases. A model for predicting one of 66 business types based only upon the business name was developed in this work (top-1 f1-score = 60.2%). Two approaches to removing the bias from this model are explored: replacing given names with a placeholder token, and augmenting the training data with gender-swapped examples. The results for these approaches is reported, and the bias in the model was reduced by hiding given names from the model. However, bias reduction was accomplished at the expense of classification performance (top-1 f1-score = 56.6%). Augmentation of the training data with gender-swapping samples proved less effective at bias reduction than the name hiding approach on the evaluated dataset. [1]

## 1 Introduction

This work presents a model that takes the text of a company name as input (e.g., "Daniel's Roofing Inc") to predict the type of the business (e.g., "Roofer"). The model is trained to remove given names from the business name in order to avoid prediction biases related to given name gender and geographic origin.

It is a common issue that a business knows the name of a small business counterparty, but does not know the archetype of the business. For example, invoice data includes business name without the business type. It is helpful for sales and marketing functions to segment clients in order to treat similar clients similarly. Small businesses present a unique challenge when performing customer segmentation, in that small businesses may not be listed in databases or classification systems for large companies that include metadata such as the company type or classification code. Beyond the sales and marketing functions, customer segmentation can also be useful for customizing service delivery to clients. The type of a company can also be useful in forecasting sales, and in many other applications. The mix of various types of companies in a client list can also reveal the trends for client segments. Finally, classifying the type of a business may expose useful equity trading signals [1] [2]. The analysis of equity trading models can often include industry classification. For example, the Global Industry Classifications Standard (GICS) [3] was applied to assess trading model performance across various industries in [4] and [5].

---

[1] Thank you to Professor Miodrag Bolic from the University of Ottawa for reviewing this manuscript

A limitation of existing industry classification systems as applied to small businesses is the lack of labels for small businesses. The criteria for inclusion in widely used industry classification systems presents a bias in the training data (i.e., the names of the companies) towards large entities (by revenue and head count). For example, a model trained on corporation names from the Russell 3000 index [6] will not be prepared properly at inference time to predict the type of business conducted at "Daniel's Barber Shoppe". Neither the business type (barber), nor the naming conventions ("Shoppe") of small businesses are reflected in the names of larger corporations such as the members of the Russell 3000.

There is, however, a potential downside to training a model to predict the type of a small business based upon the name of the business alone: introducing bias. Small business names often include gendered and geographically localized given names. This can lead the a model associating given names with business types, which can lead to regulatory fines against corporations for discrimination. For example, without adjusting for this bias, a trained model may classify "Daniel's Gems" as a Jeweller while classifying "Sandy's Gems" as a Homecraft store. Further complicating matters, out-of-vocabulary names may also be biased by confusing the classifier into picking the wrong label, or by leading to a label associated with a name-based geographic stereotype.

Having motivated the need for an unbiased model that predicts the type of a small business from the name of the business, the approach to developing such a machine learning model is now presented.

In this work, the small business classification labels utilized were derived from the city of Vancouver's Licence Bylaw 4450 [7]. The dataset used for model development was obtained from the city's open data website [8]. This dataset includes a variety of small businesses more suitable for small business classification than the commonly used industry classification systems.

The granularity of a model's predictions affects the observed performance. Specifically, consider that using a random number generator to classify a small business into one of two general categories has a 50% chance of correctly classifying, whereas the same random number generator has a 1% chance of classifying a company into one of 100 more detailed categories. Clearly, the statistical penalty for specificity justifies making a general and a specific prediction from a small business classification model. For this reason, the model presented in this work outputs a high-level prediction regarding the overall business type, along with a more specific prediction regarding the exact small business type. Top-1 and top-2 results are provided where appropriate.

The labels within the high-level small business type prediction are **B2C**, **B2B**, **PUB**, and **B2BC**. B2C stands for business to consumer, and so B2C companies sell to the consumer. Similarly, B2B companies sell to other companies on a business to business basis. B2BC companies sell to both business and consumers, and PUB is composed of government entities that are involved in delivering public services (e.g., schools, associations, and government entities). There is a mapping from the classes within the city of Vancouver's Licence Bylaw 4450 [7] to the 4 aforementioned high-level categories, as explained in later sections of this work.

A manual data audit of a sample from the dataset reveals that approximately 30% of the labels are difficult for humans to predict. Table 1 shows a representative sample of 20 records, revealing some of the limits model prediction performance.

| Business Name | True Label | Reasonable to Predict? | Potentially Classified As |
|---|---|---|---|
| Altum Spa And Wellness | Health and Beauty | YES | - |
| Paul Loubert Plumbing & Gas | Plumber | YES | - |
| Rambling Hound Bakery | Retail Dealer | YES | - |
| Teamworks Health Clinic | Health Services | YES | - |
| Tripzter Travel Inc | Travel Agent | YES | - |
| B-Lynne Clothing | Retail Dealer | YES | - |
| Fotojo Studios | Photographer | YES | - |
| Happiness Productions Inc | Production Company | YES | - |
| Kin's Farm Market | Retail Dealer - Food | YES | - |
| North Shore Neuropsychology | Health Services | YES | - |
| Grandola Contracting Ltd | Contractor | YES | Referral Services |
| New Concept Education Inc | Instruction | YES | School (private) |
| Odenza Homes Ltd | Contractor | YES | Residential/Commercial |
| XDM Digital Media | Photographer | YES | Computer Services |
| Ayreborn Audio-Video Inc | Contractor | MAYBE | Computer Services |
| Zwada Design Inc | Retail Dealer | MAYBE | Homecraft |
| E Leaders Consulting | Referral Services | NO | Contractor |
| Gerald Victor Gionco (Gerald Gionco) | Exhibitions/Shows/Concerts | NO | Contractor |
| Hangar 18 Creative Group Inc | Computer Services | NO | Studio |
| RHIQS Holdings Ltd | Laboratory | NO | Residential/Commercial |

**Table 1.** Processed random sample from dataset, indicating that some labels are difficult to correctly predict better than chance.

The "small business type classification" task is very challenging when the only information available is the company name. Some company names involve a one-of-a-kind play on words, street addresses, or are not descriptive. Others are either overly generic or overly specific. What is a reasonable performance expectation for this task? Due to the aforementioned, perhaps unsolvable, edge cases, the utility of the classifier to be "good enough" to apply as a customer segmentation signal is quite subjective. This work simply states the results observed on the dataset, and leaves that qualitative acceptance criteria up to the reader and those who choose to incorporate the described model into their own work.

The following sections will proceed to connect this work to the prior art (Section 2), present the design and evaluation of the deep learning models developed in this work (Section 3), discuss the implications and limitations of the work (Section 4), and then concludes with a summary of the key results and a discussion of possible future work (Section 5).

## 2 Prior Art

There are several human-curated industry classification systems including Standard Industrial Classification (SIC) [9], North American Industry Classification System (NAICS) [10], Global Industry Classification Standard (GICS) [3], Industry Classification Benchmark (ICB) [11], and others. One comparisons of these classification schemes revealed that "the GICS advantage is consistent from year to year and is most pronounced among large firms" [12]. This further reinforces the point that these classification systems are better at categorizing large firms, rather than small businesses.

The labels within industry classification systems provide an opportunity for supervised learning. A complimentary approach is to apply unsupervised learning to model the data, and then cluster the embedded vectors within the unsupervised model. In the case where only a short company name is available to describe the target, unsupervised learning is presented with too few factors to bridge the gap from company name to company type. However, clustering for industry classification has been applied in other work [13].

Measuring bias in a text classifier requires a validation dataset that is not a subset or split of the original dataset. The second dataset is required because without it, the testing and training data will likely contain the same data distribution, hiding bias from the assessment [14]. The purpose of the second dataset is to evaluate the classifier in out-of-distribution conditions it was not specifically trained for.

Gender bias in word embedding models could be reduced by augmenting the training data with gender swapping samples [15], fine-tuning bias with a larger less biased dataset [16] or by other weight adjustment techniques [17], or removing gender-specific factors from the model [18]. Applying several of these techniques in combination may yield the best results [16]. In this work, augmenting training with gender swapping samples, and hiding gender-specific factors from the model were evaluated.

Machine learning models are typically assessed using a top-k (also called top-n) accuracy, to show how precision and recall are affected by the specificity of the response [19]. In top-1 evaluation, an inference is counted as correct only if the highest probability label in the model output matches the true label. Similarly, in top-2 evaluation, an inference is counted as correct if either of the 2 highest probability labels in the model output match the true label. The looser constraint typically leads to higher recall and poorer precision.

## 3 Data Science and Model Development

### 3.1 Dataset Preparation

The dataset preparation began with clarifying the business name and type for each business in the dataset. To select the most descriptive name from the BusinessName and BusinessTradeName fields in the dataset, the BusinessTradeName was used as the business name unless that field was blank, in which case the BusinessName field was used as the business name. Next, business names beginning and ending with round braces (indicating the name of the business is a person's name) were removed from the dataset. The business type was extracted from the BusinessType field of the dataset. Next, duplicates from the set of business name and type pairs were dropped. These duplicates may exist because of events such as license renewals. Next, items with low frequency business type labels were dropped from the dataset. Specifically, labels with less than 100 samples were dropped.

The general category "Office" was dropped from the dataset as it was not descriptive for the purposes of building a business name classification model, and seems to be a catchall. Similar categories were merged into broader class labels when the business names within the original categories were deemed too similar to separate. Specifically, several labels from the real-estate sector (Apartment House, Pre-1956 Dwelling, Non-profit Housing, Apartment House Strata, Secondary Suite - Permanent, Multiple Dwelling, Duplex, and One-Family Dwelling) were replaced with the general label Residential/Commercial. The labels Temp Liquor Licence Amendment, Liquor Establishment Standard, Liquor Establishment Extended, Liquor License Application, and Liquor Retail Store were replaced with the more general label Liquor Establishment. Items in the category U-Brew/U-Vin were merged with the Liquor Equipment class label. The three labels Laundry-Coin Operated Services, Laundry Depot, and Laundry (w/equipment) were replaced with the more general label Laundry. The labels Ltd Service Food Establishment, Restaurant Class 1, and Restaurant Class 2 were replaced with the more general label Restaurant. Labels Short-Term Rental, and Motel were combined with the category Hotel. Contractor - Special Trades was combined with the category Contractor. Although not all business and trade schools are private schools, the labels School (Business & Trade) were merged into the category School (Private), as the names were similar in both categories. Finally, The label Artist Live/Work Studio was merged into the category Studio.

Having completed the pre-processing, the 66 business types tracked in the dataset are presented in Table 2 along with the number of samples per label.

| Business Type | Samples | Business Type | Samples |
|---|---|---|---|
| Residential/Commercial | 12145 | Manufacturer - Food | 416 |
| Contractor | 9438 | Gas Contractor | 415 |
| Retail Dealer | 5667 | Moving/Transfer Service | 414 |
| Restaurant | 5253 | Auto Repairs | 378 |
| Health Services | 3671 | Educational | 372 |
| Computer Services | 2230 | Homecraft | 361 |
| Electrical Contractor | 2146 | Caterer | 353 |
| Wholesale Dealer | 1833 | Rentals | 339 |
| Retail Dealer - Food | 1730 | Animal Services | 320 |
| Health and Beauty | 1686 | Photographer | 319 |
| Financial Services | 1487 | Electrical-Security Alarm Installation | 318 |
| Instruction | 1148 | Fitness Centre | 312 |
| Janitorial Services | 1082 | Physical Therapist | 308 |
| Plumber & Gas Contractor | 958 | Beauty Services | 264 |
| Production Company | 940 | Auto Parking Lot/Parkade | 263 |
| Community Association | 892 | School (Private) | 262 |
| Landscape Gardener | 854 | Secondhand Dealer | 254 |
| Manufacturer | 807 | Therapeutic Touch Technique | 249 |
| Personal Services | 790 | Security Services | 224 |
| Repair/Service/Maintenance | 764 | Employment Agency | 220 |
| Liquor Establishment | 750 | Printing Services | 197 |
| Plumber | 728 | Auto Dealer | 188 |
| Massage Therapist | 642 | Roofer | 157 |
| Electrical-Temporary (Filming) | 593 | Money Services | 152 |
| Cosmetologist | 591 | Scavenging | 149 |
| Entertainment Services | 565 | Laboratory | 146 |
| Real Estate Dealer | 511 | Hotel | 146 |
| Referral Services | 501 | Sprinkler Contractor | 145 |
| Painter | 478 | Jeweller | 139 |
| Exhibitions/Shows/Concerts | 449 | ESL Instruction | 127 |
| Studio | 440 | Seamstress/Tailor | 118 |
| Travel Agent | 425 | Auto Detailing | 112 |
| Wholesale Dealer - Food | 423 | Tattoo Parlour | 109 |

Table 2. Business types and their sample counts after data pre-processing

### 3.2 Initial Model with Gender and Name Origin Biases

After the pre-processing described in Section 3.1, a FastText supervised learning model was trained on the dataset with a random weight initialization [20]. A hyperparameter search of learning rates and number of training iterations resulted in the selection of 6 training iterations as the early stopping point, and a learning rate of 0.2 being selected. The embedding dimension width of the model was 100

| Business Type | Samples | Source Columns |
|---|---|---|
| B2BC | 39943 | Auto Dealer, Caterer, Computer Services, Contractor, Electrical Contractor, Electrical-Security Alarm Installation, Employment Agency, Entertainment Services, Exhibitions/Shows/Concerts, Financial Services, Gas Contractor, Hotel, Instruction, Landscape Gardener, Money Services, Moving/Transfer Service, Painter, Photographer, Plumber, Plumber & Gas Contractor, Printing Services, Production Company, Real Estate Dealer, Referral Services, Rentals, Repair/ Service/Maintenance, Residential/Commercial, Roofer, Scavenging, Security Services, Sprinkler Contractor, Studio, Travel Agent |
| B2C | 24094 | Animal Services, Auto Detailing, Auto Parking Lot/Parkade, Auto Repairs, Beauty Services, Cosmetologist, ESL Instruction, Fitness Centre, Health Services, Health and Beauty, Homecraft, Jeweller, Liquor Establishment, Massage Therapist, Personal Services, Physical Therapist, Restaurant, Retail Dealer, Retail Dealer - Food, Seamstress/Tailor, Secondhand Dealer, Tattoo Parlour, Therapeutic Touch Technique |
| B2B | 5300 | Janitorial Services, Laboratory, Manufacturer, Manufacturer - Food, Electrical-Temporary (Filming), Wholesale Dealer, Wholesale Dealer - Food |
| PUB | 1526 | Community Association, Educational, School (Private) |

**Table 3.** High-level business types, their sample counts, and the business types mapped to each high-level class after data pre-processing. An sample imbalance between the classes can be observed.

dimensions, and the window size was set to 5. Model performance is recorded in Table 4.

This initial model was trained on business names from one geographic part of the world (Western Canada). The dataset included a variety of local given names and family names in the business name training data. The dataset is from an English language speaking province. These factors represent a strong potential for classification bias based upon the gender of a given name within a business name, and the geographic origin of names within business names.

Two approaches were applied to assess model bias. In the first approach, the model was assessed by constructing out-of-distribution data composed of a randomly selected given name followed by a random dictionary word with the first letter capitalized. For example, "Daniel's Mirror". In the second approach, text from the model's testing dataset (data held back from the model training) was appended to a randomly selected person's name. For example, "Daniel's Bob's Grocery Store". In order to examine bias within the model, the generation process controlled for the gender and geographic origin of the person's name for both of these approaches. The lists of names by gender and geographic origin are shown in Table 5.

| Model | Classes | Top-1 Precision (%) | | Top-1 Recall (%) | | Top-2 Precision (%) | | Top-2 Recall (%) | |
|---|---|---|---|---|---|---|---|---|---|
| | | $\mu$ | $\sigma$ | $\mu$ | $\sigma$ | $\mu$ | $\sigma$ | $\mu$ | $\sigma$ |
| Initial Model (Section 3.2) | All 66 | 60.2 | 0.4 | 60.2 | 0.4 | 36.7 | 0.3 | 73.3 | 0.6 |
| Initial Model | B2B | 49.5 | 1.5 | 40.6 | 2.1 | - | - | - | - |
| Initial Model | B2BC | 85.9 | 0.6 | 85.2 | 0.8 | - | - | - | - |
| Initial Model | B2C | 75.2 | 0.8 | 79.2 | 0.8 | - | - | - | - |
| Initial Model | PUB | 55.6 | 4.0 | 51.5 | 3.8 | - | - | - | - |
| Name Replacement (Section 3.3) | All 66 | 56.6 | 0.7 | 56.6 | 0.7 | 35.1 | 0.3 | 70.2 | 0.6 |
| Name Replacement | B2B | 41.6 | 1.3 | 34.7 | 2.0 | - | - | - | - |
| Name Replacement | B2BC | 84.5 | 0.5 | 81.6 | 0.7 | - | - | - | - |
| Name Replacement | B2C | 70.7 | 0.7 | 77.3 | 0.8 | - | - | - | - |
| Name Replacement | PUB | 49.3 | 3.1 | 45.1 | 3.7 | - | - | - | - |
| Training Augmentation (Section 3.4) | All 66 | 57.9 | 0.6 | 57.9 | 0.6 | 35.5 | 0.2 | 71.0 | 0.5 |
| Training Augmentation | B2B | 43.0 | 1.9 | 37.0 | 2.0 | - | - | - | - |
| Training Augmentation | B2BC | 84.8 | 0.9 | 84.0 | 0.7 | - | - | - | - |
| Training Augmentation | B2C | 73.4 | 0.7 | 77.1 | 0.7 | - | - | - | - |
| Training Augmentation | PUB | 50.8 | 5.0 | 48.1 | 4.9 | - | - | - | - |

**Table 4.** Classification performance report. These results are an average of 10 training and testing runs (testing n=7,087). Predictions involving all 66 classes were assessed for top-1 and top-2 performance, whereas only top-1 performance is reported when predicting one of the four high-level labels.

### 3.3 Replacing Names in Training Data and at Inference Time

To address the identified biases in the initial model, the training data was further processed to replace with a token given names from within the company names data. The given name replacement task was first attempted using spaCy's named entity recognition capability [23], but ultimately a dictionary-based approach proved more effective for the dataset in question. To obtain the results reported in this work, a long list of male and female names was obtained from a python library [24], which in turn obtained the name list from the 1990 U.S. Census [25]. One issue with removing all names from the business name is that a business name may be deleted if the full name of the business is a name, causing classification to fail. For example, the name "Denny's" simply disappears into a blank string. Another case where this issue arises is businesses registered as a person's name e.g., "John Smith". To avoid this problem, a standardised string was used to represent the replaced name. Specifically, the character "_" was used as the replacement token. The key insight with this approach is that instead of trying to learn about out-of-distribution "foreign" names or gendered names, the model can learn less about local given names and their related gender.

After this additional pre-processing, a new FastText supervised learning model was trained on the dataset. A hyperparameter search of learning rates and number of training iterations resulted in the selection of 6 training itera-tions as the early stopping point, and a learning rate of 0.2 being selected. The embedding dimension width of the model was 100 dimensions, and the window size was set to 5. The performance of this model is recorded in Table 4.

| Names List Type | Names List | List Origin |
|---|---|---|
| Canadian Female | Olivia, Emma, Charlotte, Sophia, Aria, Ava, Chloe, Zoey, Abigail, Amilia | [21] |
| Canadian Male | Noah, Liam, Jackson, Lucas, Logan, Benjamin, Jacob, William, Oliver, James | [21] |
| Mexican Female | Ximena, Valentina, María, Fernanda, Sofía, María, José, Martina, Emilia, Zoe, Mia, Dulce | [22] |
| Mexican Male | Santiago, Mateo, Matías, Diego, Sebastián, Nicolás, Miguel, Ángel, Iker, Alejandro, Samuel | [22] |

Table 5. Lists of names used for model evaluation

| Model Evaluation | Given Name Data Used | Imbalanced classification of business names (%) | | | | | |
|---|---|---|---|---|---|---|---|
| | | Initial Model | | Replace Names | | Augment Training | |
| | | $\mu$ | $\sigma$ | $\mu$ | $\sigma$ | $\mu$ | $\sigma$ |
| **Approach 1:** Random given name + word (Out-of-distribution) | Canadian male + male | 0.27 | 0.04 | 0.00 | 0.00 | 0.00 | 0.00 |
| | Canadian female + female | 1.08 | 0.29 | 0.00 | 0.00 | 0.07 | 0.01 |
| | Canadian male + female | 9.85 | 2.66 | 0.00 | 0.00 | 10.12 | 0.03 |
| | Canadian male + Mexican male | 18.88 | 3.82 | 0.02 | 0.00 | 0.00 | 0.00 |
| | Canadian female + Mexican female | 18.78 | 3.89 | 0.04 | 0.02 | 9.74 | 0.02 |
| | Canadian male + Mexican female | 19.26 | 2.65 | 0.00 | 0.00 | 0.00 | 0.00 |
| Average | | 11.35 | 2.23 | 0.01 | 0.00 | 3.32 | 0.01 |
| **Approach 2:** Random given name + model testing data (True dist.) | Canadian male + male | 0.90 | 0.02 | 0.00 | 0.00 | 0.00 | 0.00 |
| | Canadian female + female | 0.65 | 0.02 | 0.31 | 0.03 | 0.92 | 0.04 |
| | Canadian male + female | 2.04 | 0.08 | 0.90 | 0.08 | 5.60 | 0.54 |
| | Canadian male + Mexican male | 2.08 | 0.09 | 0.72 | 0.11 | 0.00 | 0.00 |
| | Canadian female + Mexican female | 1.68 | 0.17 | 0.57 | 0.08 | 4.88 | 0.55 |
| | Canadian male + Mexican female | 1.87 | 0.08 | 0.31 | 0.05 | 0.00 | 0.00 |
| Average | | 1.54 | 0.08 | 0.47 | 0.06 | 1.90 | 0.19 |

Table 6. Model Bias Evaluation: Prediction disagreements based upon gender and geographic origin of given names. For the rows of Approach 1, 10,000 samples were compared to 10,000 other samples. For the rows of Approach 2, 7,087 samples were compared to 7,087 other samples using the model training data as a basis. Note: To focus on large-scale bias, imbalances of less than 5 samples within a class were not considered. Each prediction was one of 66 possible labels.

The observed drop of approximately 4% in top-1 precision and recall relative to the original model is an understandable outcome of removing bias within the dataset which was providing signal for classification. This drop indicates that perhaps some classification performance was originating in name-based biases.

### 3.4 Augmenting Training Data with Additional Gender Information

A second method for addressing the identified biases in the initial model is to augment the training data with gender-swapped copies of the text. For example, the business name "Alice and Associates Plumbing Ltd" with the label Plumber can be observed as containing a given name within the aforementioned list of female given names, and a new training record with the label Plumber can be added to the dataset: "Bob and Associates Plumbing Ltd". Similarly, training records containing male given names can be augmented with a gender swapped version. The intuition is that a balance of gender information per label could cancel out gender bias for each label.

After the training data augmentation, a new FastText supervised learning model was trained on the dataset. A hyperparameter search of learning rates and number of training iterations resulted in the selection of 6 training iterations as the early stopping point, and a learning rate of 0.2 being selected. The embedding dimension width of the model was 100 dimensions, and the window size was set to 5. The model performance is recorded in Table 4.

## 4 Discussion

Observe in Table 6 that as expected, the model evaluation on the out-of-distribution dataset (Approach 1) revealed a much higher bias than observed when testing on the within-distribution dataset (Approach 2). The approach to replace names with a static string during training and inference resulted in the elimination of bias in the out-of-distribution test. Specifically, in Approach 1 the classification imbalance per class dropped from 11.35% ($\sigma = 2.23\%$) to 0.01% ($\sigma = 0.00\%$). Observing the results on the test data held back from the training data of the company names dataset, observe that the bias dropped from 1.54% ($\sigma = 0.08\%$) down to about a third, at 0.47% ($\sigma = 0.06\%$). The significant drop in observed bias for both evaluation approaches motivates the publication of this work.

Surprisingly, augmentation of the training data with gender swapping samples worked in [15], but was not as effective on the dataset studied in this work. The out-of-distribution bias results (Approach 1) were 3.32% ($\sigma = 0.01\%$), which is significantly lower than the original model. However, the results were not as strong as the names replacement approach, and came at the expense of lower model performance and a worse imbalance on the test data from the company names dataset 1.90% ($\sigma = 0.19$).

Model stability was an issue with both the name replacement and data augmentation with gender swapping approaches. The models were sensitive to very small changes. For example, models were sensitive to the capitalization of words in the business name such that "Bob's Plumbing" was classified as Plumber, while "Bob's plumbing" was classified as Restaurant. This sensitivity was a major factor in the classification disagreements between the samples fed to the models. For example, in the name replacement model, the input samples "Aria Lodge & Associates _ Ltd" and "_ Lodge & Associates _ Ltd" resulted in different class predictions, even though the text only differs by a few characters.

Note that although the results imply that replacing names perfectly replaces all names in the training and testing data, this is not the case. The predictions do converge to result in very low classification disagreement based upon given names, but the dataset itself still includes several names that were not picked up by the 1990 U.S. Census names list. For example, the name Aria from the Canadian Female names list, and the name Ximena from the Mexican Female names list were both not removed by the name replacement approach. Furthermore, many names within the dataset such as Ho and Fraser were not removed. This observation implies that removing most given names from the training data was sufficient. Also, there were scenarios where a token within the text was incorrectly replaced, deleting some information that could have been used for classification. For example, the business name "Lodge & Associates Investigations Ltd" lost the word "Investigations" because it has a substring "In" that matches a name in the names list. Removing this one case ("In") from the names list did not improve model performance significantly, and so there is likely a collection of such name replacement precision and recall improvements that collectively would improve the model's overall precision and recall. This additional direction for improvement is left as future work.

## 5 Conclusion and Future Work

In this work, a small business type classifier with gender and geographic origin biases was developed (f1-score = 60.2%). Although the bias in the model was reduced by hiding given names from the model, this was accomplished at the expense of model performance (f1-score = 56.6%). Augmentation of the training data with gender swapping samples proved less effective than the name hiding approach. In future work, the precision and recall of the name hiding method employed in this work may be improved.